\documentclass[journal]{IEEEtran}

\usepackage{array}
\usepackage{graphicx}
\usepackage{stfloats}
\usepackage{amssymb}
\usepackage{caption}
\usepackage{color, colortbl}
\usepackage{bm}
\usepackage{subcaption}
\usepackage{todonotes}
\definecolor{Gray}{gray}{0.8}

\graphicspath{{figures/}}

\begin{document}
\title{Deep Network for Scatterer Distribution Estimation for Ultrasound Image Simulation}

\author{Lin~Zhang,
        Valery~Vishnevskiy,
        and~Orcun~Goksel
\thanks{L.~Zhang, V.~Vishnevskiy, and O.~Goksel are with the Computer-assisted Applications in Medicine group, ETH Zurich, Switzerland.}
\thanks{Funding provided by the Swiss Innovation Agency Innosuisse \#25540PFLS.}
}

\markboth{Zhang \MakeLowercase{\textit{et al.}}: Scatterer Distribution Estimation}
{Zhang \MakeLowercase{\textit{et al.}}: Scatterer Distribution Estimation}

\maketitle

\begin{abstract}
Simulation-based ultrasound training can be an essential educational tool.
Realistic ultrasound image appearance with typical speckle texture can be modeled as convolution of a point spread function with point scatterers representing tissue microstructure. 
Such scatterer distribution, however, is in general not known and its estimation for a given tissue type is fundamentally an ill-posed inverse problem.
In this paper, we demonstrate a convolutional neural network approach for probabilistic scatterer estimation from observed ultrasound data.
We herein propose to impose a known statistical distribution on scatterers and learn the mapping between ultrasound image and distribution parameter map by training a convolutional neural network on synthetic images.
In comparison with several existing approaches, we demonstrate in numerical simulations and with in-vivo images that the synthesized images from scatterer representations estimated with our approach closely match the observations with varying acquisition parameters such as compression and rotation of the imaged domain.
\end{abstract}

\begin{IEEEkeywords}
Deep learning, image synthesis, deconvolution, medical training
\end{IEEEkeywords}

\IEEEpeerreviewmaketitle

\section{Introduction}

\IEEEPARstart{U}{ltrasound} (US) is a low-cost, non-invasive and portable imaging modality, and it has been widely used in the clinical routine, especially for obstetrics and gynaecology as it does not involve ionizing radiation.
However, ultrasound scanners produce images suffering from limited spatial resolution, signal-to-noise ratio and tissue contrast, which make the interpretation of images very difficult. 
Clinical ultrasound practice relies exclusively on sonographer expertise in navigating a hand-held probe and visual inspection of the acquired images; and an extensive training is required for being able to conduct such clinical examinations. 
The current education is mostly based on training on real patients, under the guidance of expert specialists. 
This form of education is inefficient due to time consumption of both specialists and patients and involve difficulties in finding volunteers especially when rare pathologies are concerned. 
Indeed, after completing the one year education students will only have seen up to 80\% of possible pathologies~\cite{kohn2004ves}. 
Learning from real ultrasound images offline for training is an alternative option that does not require volunteers. However, this approach does not allow interactive training and is highly restricted to the available clinical images.
Computer-assisted ultrasound simulation can aid training, especially important for rare pathologies, which are still vital to identify.
New scenarios of any given case can be simulated with transformations on pre-collected data, and different imaging parameters and conditions can be simulated from the same data.
Furthermore, computerized simulations would allow for training in a real-time virtual-reality environment with complex anatomical scenarios and various pathologies, from which the medical students can obtain in-depth knowledge of possible clinical scenarios during training.

Wave propagation, convolution and ray-based methods are the three major techniques for US simulation. 
Wave-based approaches~\cite{verweij2014, treeby2012modeling} model ultrasound propagation in tissue by solving complex acoustic wave equations.
Therefore, these approaches are relatively slow and not suitable for real-time applications. 
Convolution-based~\cite{jensen1996field} methods approximate the ultrasound interactions as a spatial impulse response (point-spread function, PSF), which is then convolved with a representation of sub-wavelength tissue structures, known as \emph{scatterers}, under the assumption that the acoustic field is linear.
This can realistically reproduce typical ultrasound noisy texture, known as speckle, caused by the constructive and destructive inference of echoes scattered by countless tissue scatterers. 
On the one hand, speckle can be seen as noise that degrades tissue contrast~\cite{tay2010ultrasound}, but on the other hand, it can help distinguish tissues and identify pathologies~\cite{alessandrini2011restoration}. 
Restoration of this granular pattern is not only important for visual realism in a simulation scenario, but also for preserving structural and diagnostic information about the tissue.
Ray-based methods~\cite{burger2012real,mattausch2018realistic} simulate the propagation of ultrasonic wavefront as rays using computed graphics techniques, which allows to simulate interactions such as refractions and reflections, while simulating speckle using a PSF convolution with a texture representing scatterers.
Using stochastic and sophisticated interaction models with Monte-Carlo ray sampling, this was shown in~\cite{mattausch2018realistic} to lead to impressively realistic US images even at real-time framerates.

There are recent works that use deep learning, and in particular generative adversarial networks, for ultrasound image synthesis~\cite{hu2017freehand, tom2018simulating}.
Nevertheless, these works aim to synthesize isolated individual images, without any intermediate (physical space) representation or parametrization, they are not designed for simulating images with speckle motion coherent with underlying physical motion, and therefore may not be applicable in a real-time, interactive simulation framework where temporal continuity of simulated frames is utmost important.
In other words, for an infinitesimal movement of the probe to one side should result in the image moving infinitesimally to the other side.
If every image is generated somewhat independently, this behaviour and the visual fluidity cannot be guaranteed.
Whereas, using a fixed scatterer representation, the above condition can be satisfied.
Such tissue representations can also allow scene editing operations such as copying and adding anatomy in the scatterer domain avoids any image artifacts, as shown in~\cite{mattausch2017image}.

The problem with using scatterers in simulations is that such tissue representations are not known a priori.
Assuming it can be modeled as a PSF convolution, finding a representation from observed US data can then be posed as a blind deconvolution problem.
There has been several approaches and approximations to this problem, including inverse problem based solutions~\cite{chen2015compressive}, variational methods~\cite{zhao2016joint, alessandrini2011expectation}, and filtering techniques~\cite{michailovich2007blind, taxt1995restoration}. 
Different from our motivation of ultrasound simulation, however, the above methods aim for image denoising / restoration to achieve higher contrast or quality for better diagnostic information.
An iterative solution to the regularized deconvolution problem was proposed in~\cite{mattausch2017image} by solving the inverse problem jointly for multiple acquisitions of the same tissue, which are obtained efficiently with electronic beam steering.
This allows to overconstrain the problem with observations from multiple PSFs.
Discrete scatterer reconstructions were performed on a fine Cartesian grid in order to approximate sub-wavelength particles, where sparsity was enforced by using an $\ell_1$-norm regularization.
However, solving such inverse problem is computationally and memory-wise only possible for small patches, which were tiled in~\cite{mattausch2017image} over the imaging field, and still required hours to days to solve for a single image.
Accordingly, such a method is not efficient and scalable for scans of many images or to apply in 3D.

Scatterers can alternatively be estimated as statistical distributions of random variables.
A pipeline for generating synthetic echocardiographic ultrasound sequences was presented in~\cite{alessandrini2015pipeline}. 
This was later extended to simulate vendor specific US images for speckle tracking algorithms~\cite{alessandrini2017realistic}.
For this purpose, a scatterer space was populated with randomly sampled 3D cloud points, whose amplitudes were assigned according to the template B-Mode images after the compensation for the log-compression.
Simulated speckle statistics were reported to be in good agreement with the known fitting distributions.
This approach however does not take the point spread function, and therefore the constructive and destructive interference between scatterers into account, and sampling from B-mode (similarly from RF envelope) assumes scatterers all contributing non-negatively, which is not the case for modulated RF nature of typical US PSF. 
A simple Gaussian-parametrized model was fit to the inverse-problem reconstructed scatterers in~\cite{mattausch2015scatterer}.
This method was demonstrated for homogeneous tissues that other instances of the same tissue can be obtained with the found model for simulating new images, which are however reported as lack of visual variety of real tissues in~\cite{mattausch2018realistic}. 

For a given statistical model for scatterer distribution, we propose herein to estimate distribution parameter maps directly from observed ultrasound images.
This can be used to instantiate new scatterer maps that would reproduce the original images when input to a convolution-based simulation.
As the estimated parameter maps would represent a physical tissue space, they could be used to simulate new images faithfully with varying imaging conditions, viewing directions, and other imaging parameter variations.
Due to the power of deep neural networks in learning patterns of visual inputs, we propose to learn the mapping between simulated images and parameter maps by training a convolutional neural network.

\section{Background}
\subsection{Forward Problem of Ultrasound Simulation}
Based on the first order Born approximation (weak scattering) for soft tissues~\cite{jensen1993deconvolution}, the interaction between ultrasound field and tissue scatterers can be formulated as a 2D convolution model in discrete domain:
\begin{equation}
    \mathbf{I}[l,a] = \mathbf{g}[l,a]*\mathbf{h}[l,a] + \bm{\gamma}[l,a],
\label{eq:forward_model}
\end{equation}
with radio-frequency (RF) US image intensity $\mathbf{I}[l,a]$, scatterer intensity $\mathbf{g}[l,a]$, spatial variant point spread function (PSF) $\mathbf{h}[l,a]$ and noise term $\bm{\gamma}[l,a]$. 
$[l,a]$ are the lateral and axial coordinates with respect to the probe origin, elevational thickness is ignored here. 

Ultrasound point spread function can be approximated with a two dimensional Gaussian kernel modulated by a cosine function in the axial direction~\cite{burger2012real}, i.e.:
\begin{equation}
    \mathbf{h}[l,a] = e^{-\frac{l^2}{\sigma_l^2}-\frac{a^2}{\sigma_a^2}} \cos(2\pi f_c a),
\label{eq:psf}
\end{equation}
where $f_c$ is the transducer center frequency, $\sigma_l$ and $\sigma_a$ determines the Gaussian shape along the lateral and axial direction.
Due to nonuniform focusing and aperture, the PSF in US imaging is often assumed to be spatial-variant mainly along the axial direction~\cite{nagy1998restoring, mattausch2017image, alessandrini2011restoration}.
Some recent works model PSF as continuously non-stationary blurring, such as based on semigroup theory~\cite{michailovich2017non} and with the diffraction effects during wave propagation~\cite{besson2019physical}. 
For deconvolution tasks, PSF has often been assumed to be patchwise invariant, an approach we also adopt in this paper.

\subsection{Inverse Problem of Scatterer Reconstruction}
Eq.~(\ref{eq:forward_model}) can be equivalently written in a matrix-vector form as $\mathbf{Ax} + \mathbf{n} = \mathbf{b}$ with the convolutional matrix $\mathbf{A} \in \mathbb{R}^{M \times N}$ associated with PSF, a vector of scatterer amplitudes $\mathbf{x} \in \mathbb{R}^{N}$, RF image intensities $\mathbf{b} \in \mathbb{R}^{M}$ and the acquisition noise term $\mathbf{n} \in \mathbb{R}^{M}$. 
Assuming that $N=M$ and imposing no constraints on $\mathbf{x}$, a solution of this system of linear equations is referred to as the tissue reflectivity function (TRF)~\cite{taxt1995restoration, zhao2016joint}.
To mitigate ill-posed nature of this deconvolution problem, regularization is typically imposed to introduce a prior knowledge about the solution. 
Wiener filter~\cite{hundt1980digital} is a common choice for efficient image deconvolution, which solves the inverse problem based on $\ell_2$-norm regularization of the solution magnitude.
Other regularizations have been also widely explored, such as a $\ell_1$ or $\ell_p$-norm based on the assumption of Laplacian~\cite{michailovich2007blind} and generalized Gaussian distribution~\cite{alessandrini2011restoration} for the TRF $\mathbf{x}$.
Several computationally efficient methods have been proposed for TRF deconvolution with sophisticated forward models, such as axially varying kernels~\cite{florea2018axially}, physical model accounting for diffraction effects~\cite{besson2019physical}.
However, such TRF representation is difficult to attribute to a physical quantity and without any constraints on $\mathbf{x}$, the Wiener filter solution may overfit to the observation $\mathbf{b}$, e.g. a slight change in acquisition parameters may yield largely different TRF estimates.

For image simulation purpose, estimated discrete scatterer map $\mathbf{x}$ needs to have sufficiently fine resolution to approximate the underlying continuum of scatterers, hence $N\gg M$. 
Mauttausch et al.~\cite{mattausch2017image} used the $\ell_1$ norm, favoring sparse scatterer map.
Assuming a Laplacian noise distribution for $n$, the objective is formulated as follows:
\begin{equation}
    \mathbf{\hat{x}} = \arg \min_\mathbf{x} ||\mathbf{Ax}-\mathbf{b}||_1 + \lambda||\mathbf{x}||_1, \quad \textrm{s.t.} \, \mathbf{x}\geq 0,
\end{equation}
with a regularization parameter $\lambda$. 
This formulation, known as regularized least absolute deviations (RLAD), allows more robust solution with respect to outliers in the error function, which could be caused by wave interactions other than scattering, e.g. directional reflections. 

\section{Methods} \label{Sec: methods}
For US simulation, we aim to find a tissue representation from observed image, which can be used to simulate the same tissue with varying imaging conditions. 
Rather than estimating a deterministic scatterer locations and amplitudes by solving a large-scale inverse problem, we proposed to impose a statistical model on the scatterer distribution and infer corresponding parameters from the observation. 
We learn the mapping from a single US image to its parameter map in a supervised manner. 
The required paired data are generated by simulation, since no ground truth of tissue scatterer is available.
An overview of our proposed pipeline, referred as ScatParam, can be seen in Fig.1(a).
\begin{figure*}
\centering
\minipage{0.7\textwidth}
 \centering
 \includegraphics[width=1\linewidth]{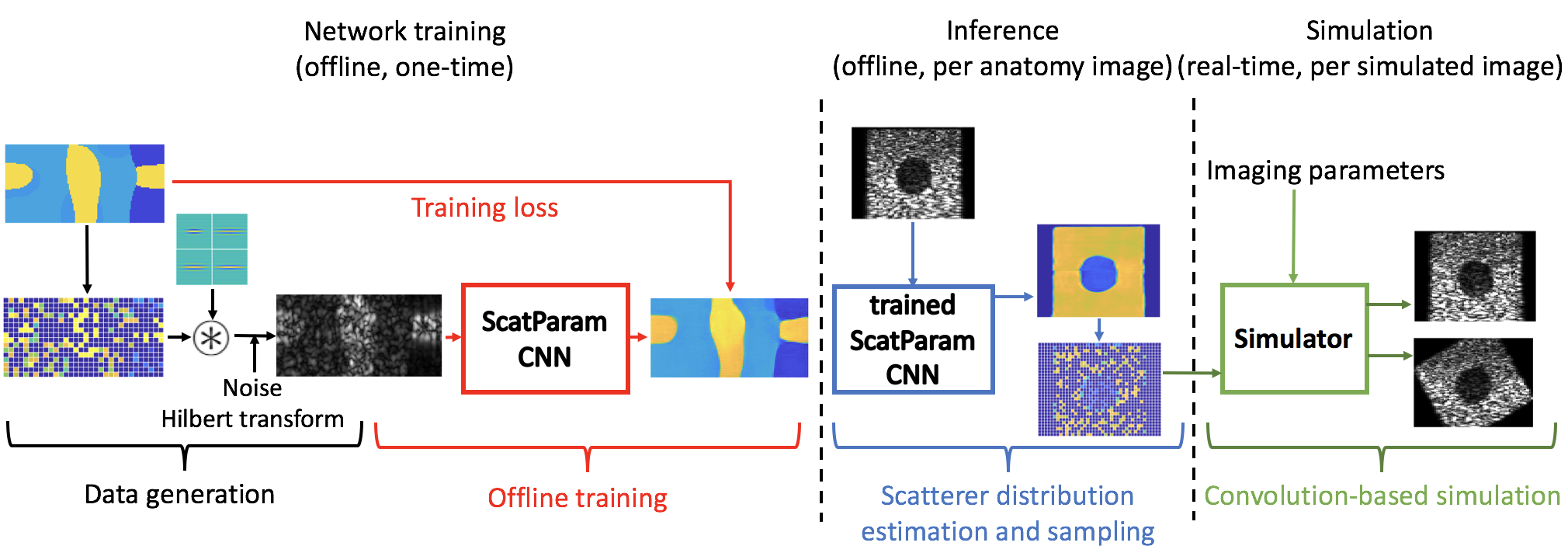}
 \caption*{(a)}
\endminipage
\hfill
\minipage{0.3\textwidth}
 \centering
 \includegraphics[width=1\linewidth]{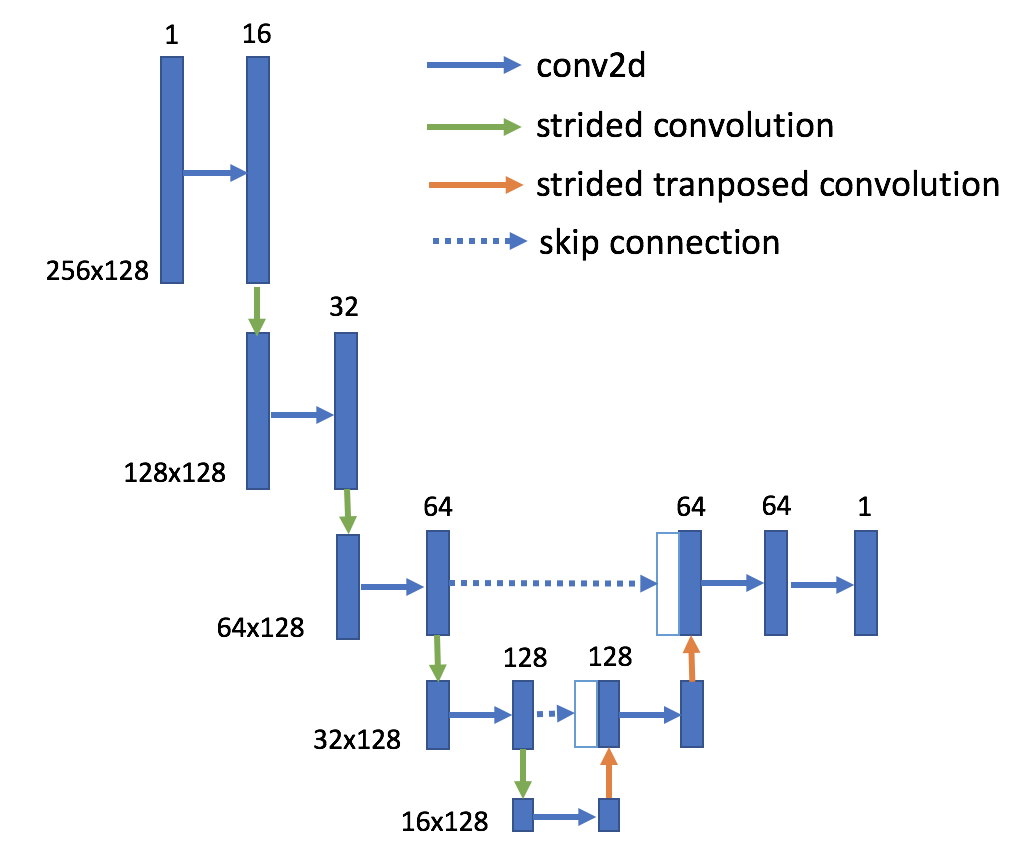}
  \caption*{(b)}
\endminipage
\caption{(a) Illustration of the proposed pipeline, note that PSF and scatterer maps are sampled for each training pair. (b) ScatParam CNN: we use stride of 2 to downsample axial dimension with image size indicated on the left for each layer, while the number of filters is shown on top of each layer.
}
\label{fig:pipeline}
\end{figure*}

\subsection{Statistical Model of Scatterer Distribution}
We assume that each tissue type can be parametrized by a model with three parameters $(\rho_s, \mu_s, \sigma_s)$, the parameter $\rho_s\in[0,1]$ for scatterer density, the mean $\mu_s$ and standard deviation $\sigma_s$ for scatterer amplitude modeled to be normally distributed.
Scatterer maps are sampled as follows: for each pixel, a Bernoulli distributed random variable is sampled, where the pixel takes the value one with probability $\rho_s$ and zero with $1-\rho_s$. For non-zero pixels, their amplitudes are sampled from a Gaussian distribution $\mathcal{N}(\mu_s, \sigma_s)$.
Tissue scattering strength is controlled by the mean, while $\sigma_s$ models random fluctuations around that mean.
Since we only consider fully developed speckles, we set $\rho_s$ to a fixed value satisfying the Rayleigh criterion; herein set to a minimum of 100 scatterers per $\textrm{mm}^2$ for a fully-developed speckle pattern~\cite{oosterveld1985texture}.

The Rayleigh criterion and distribution statistics for estimating interference with isolated scatterers~\cite{oosterveld1985texture} has only been studied for scatterers distributed randomly at continuum spatial locations.
Typical convolution based simulation packages, such as Filed\,II~\cite{jensen1996field} and SIMUS~\cite{shahriari2018meshfree}, accordingly use scatterer representations with floating-point locations in continuum domain.
Reconstructing these on a discrete map therefore necessitate a sufficiently high grid resolution to approximate the continuum.
Following~\cite{mattausch2017image} we choose to use an isotropic grid spacing (i.e.\ Cartesian grid) with the native axial resolution of the raw RF data, which inherently also satisfies the Nyquist criterion for lateral sampling. 
For instance, for a sampling frequency of 40 MHz and speed-of-sound of 1540 m/s, we use a scatterer map with a resolution of roughly $20\,\mu$m and with 5\% of the pixels populated with scatterers, resulting in a scatterer density of 130 per $\textrm{mm}^2$.

\subsection{Training Set Generation} 
\begin{figure}
\centering
\minipage{0.16\textwidth}
 \centering
 \includegraphics[width=1\linewidth]{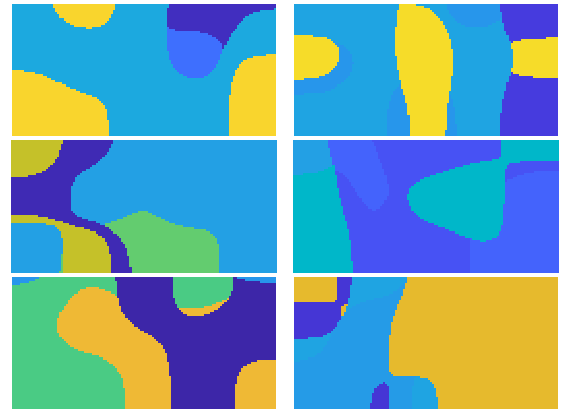}
 \caption*{(a)}
\endminipage
\minipage{0.16\textwidth}
 \centering
 \includegraphics[width=1\linewidth]{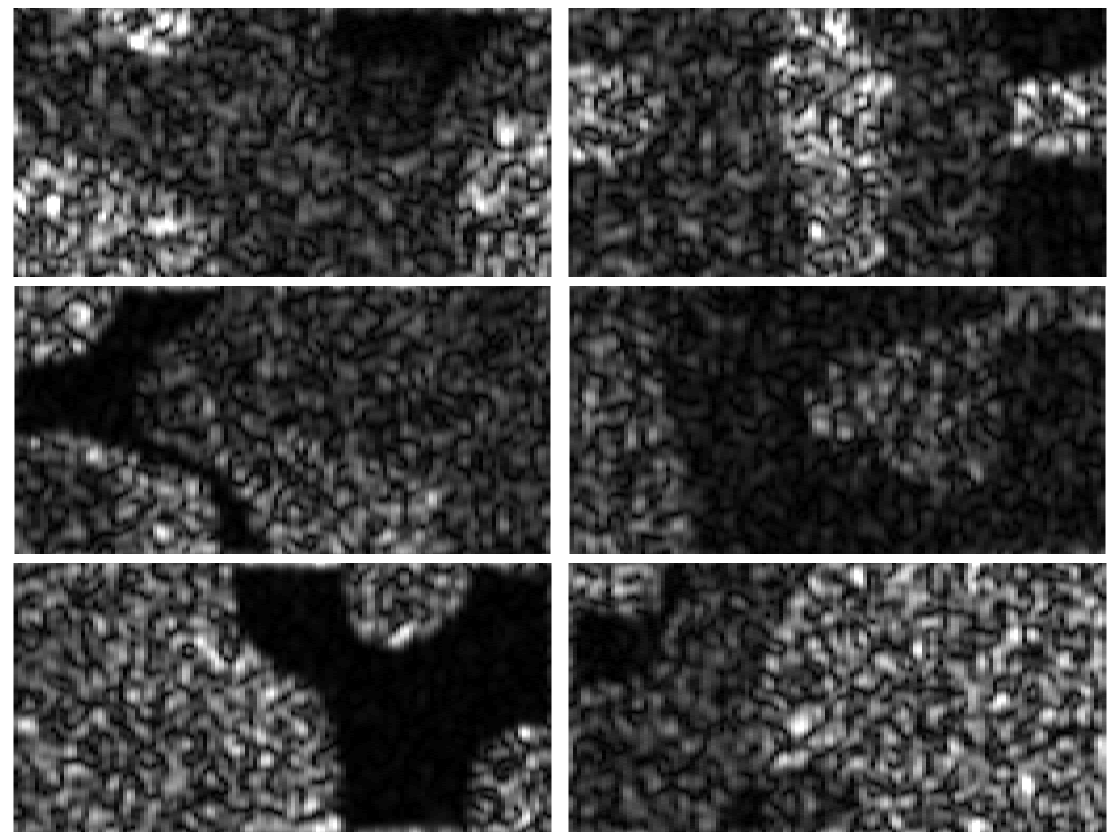}
  \caption*{(b)}
\endminipage
\minipage{0.16\textwidth}
 \centering
 \includegraphics[width=1\linewidth]{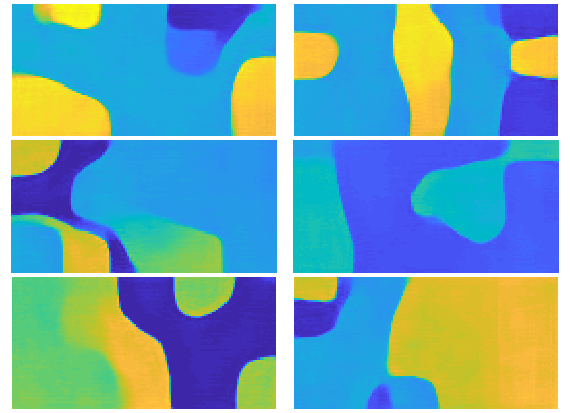}
  \caption*{(b)}
\endminipage
\caption{Generation of training images: (a)~scatterer parameter maps, (b)~envelope images simulated with scatterers sampled from the corresponding parameter maps, and (c)~the corresponding parameter maps estimated by our method. }
\label{fig:training_images}
\end{figure}

Since scatterers are an abstract tissue representation, no point-wise ground truth exists, thus we create network training data by means of simulation.
For parameter map generation, we use random synthetic shapes by overlapping irregular geometric shapes, with a procedure similar to~\cite{vishnevskiy2019deep}, where random coarse gray-scale patterns are interpolated at a finer resolution and finally thresholding them to create random shapes, as exemplified in Fig.~\ref{fig:training_images}(a).
These aim to represent a rich variety of potential tissue structures without assuming particular anatomical priors, both to be invariant to any anatomical assumptions and region-of-interest as well as to allow the network for better generalization.
We assume uniform distribution for Gaussian mean: $\mu_s \sim \mathcal{U}(0,1)$ and a fixed Gaussian standard deviation $\sigma_s=0.05$.
We assign one sampled mean value to each region, assuming the scatterers in each tissue region following the same distribution.
The scatterer maps are sampled according to the procedure described in Section~\ref{Sec: methods}(A).
RF images are generated by convolving a sampled scatterer map with a PSF.
We assume spatially invariant PSFs for image patches, allowing very fast US image generation.
PSFs are sampled from the analytic expression in Eq.~(\ref{eq:psf}), with the transducer center frequency $f_c = 6$ MHz, the sampling frequency $f_s = 40$\,MHz and normalized by its $\ell_2$ norm.
The vertical and axial spreading are uniformly sampled from $\sigma_l^2 \in [0.2,1]$ $\textrm{mm}^2$ and $\sigma_a^2 \in [0.02,0.05]$ $\textrm{mm}^2$. 
The images are then corrupted by additive Gaussian noise.
The noise level was uniformly sampled in the interval $[2, 20]\%$ of the average signal value.  
The envelope images are used as the input to the neural network, which are taken as the absolute value of Hilbert transform of RF images, illustrated in Fig.~\ref{fig:training_images}(b).
For training, 4000 parameter maps of 64$\times$128 pixels with the random geometric shapes above were first generated offline. 
Then on-the-fly during each training batch, random scatterer maps were spatially sampled from a random subset of the parameter maps, following which PSF convolution and envelope detection (Hilbert transform) were also carried out on-the-fly for each training image.

\subsection{Network Architecture and Training}
The general network architecture for ScatParam is illustrated in Fig.~\ref{fig:pipeline}(b). 
An encoder-decoder network is used to extract features from the input US image and estimate its corresponding parameter map, used for scatterer map sampling.
The model comprises an encoder and a decoder part along the axial direction, with skip connections between the corresponding layers.
This design choice is due to very low lateral resolution in US imaging being $10-20$ times lower than the axial.
Parameter maps are estimated at a coarser axial resolution than input envelope images, assuming spatially smooth tissue content. 
This facilitates a more efficient utilization of network weights and hence inference power. 
Our preliminary experiments with equally high resolution in encoder and decoder did not indicate results substantially superior to our presented architecture.
We use strided convolution to perform layer pooling and upsampling.
Exponential linear unit activation~\cite{clevert2015fast} is used at each layer except the output layer, which is linear.
The network is trained using Adam optimizer~\cite{kingma2014adam} with a learning rate of $10^{-4}$, minimizing the $\ell_1$-norm based loss function between the true $\mathbf{x}$ and estimated parameter maps $\hat{\mathbf{x}}$ given the acquisition $\mathbf{y}$ as the input:
\begin{equation}
    L(\Theta) = \mathbb{E} \|\mathbf{x} - \hat{\mathbf{x}}(\mathbf{y};\Theta)\|_1,
\end{equation}
with the network parameters $\Theta$, the empirical average $\mathbb{E}$ over the training sampling procedure.
The batch size is set to 16. 
The network is trained for 20000 iterations.

The proposed pipeline shown in Fig.~\ref{fig:pipeline} is summarized as follows:
\begin{enumerate}
	\item Data generation: simulation of synthetic ultrasound images involves scatterer map sampling and convolution with point spread functions;
	\item Offline training: a convolutional neural network is trained with the simulated paired data for  parameter map estimation;
	\item Scatterer distribution estimation and sampling: for each observation, a scatterer map is sampled from the estimated distribution parameter map for synthesizing new images;
	\item Convolution-based simulation: the sampled scatterer map is fed into the convolution based simulator, which generates images with desired imaging parameters in real time.
\end{enumerate}

\section{Experiments and Results}  \label{experiments}
We study our proposed method comparatively to its alternatives with experiments conducted on numerical simulations of synthetic phantoms, as well as on actual data acquired from a gelatin phantom and in vivo tissue.
Given the background above, we consider three alternatives to compare our method against:
\begin{itemize}
\item Sampling scatterers from envelope image ({\bf SampleEnv}): This is an adaptation of the method proposed in~\cite{alessandrini2015pipeline} for sampling continuous scatterers from log-uncompressed B-mode images.  
We herein adapt this to sample on discrete scatterer maps and, for better accuracy, we use the original envelope images instead, as we have access to them.
Background (non-myocardial) scatterer amplitudes in~\cite{alessandrini2015pipeline} were updated at each simulation step using the corresponding input image. Treating scatterers as physical tissue-embedded entities, we herein keep their amplitudes fixed while their spatial locations may change, similarly to the works on displacement tracking and elastography.
\item Tissue reflectivity function ({\bf TRF}) : This is the simple Wiener filter estimation to the deconvolution inverse problem~\cite{jensen1994nonparametric, taxt1995restoration}.
We use a spatially constant filter kernel, as the PSF computed or estimated at the center of the imaged field of view.
\item Iterative scatterer reconstruction ({\bf ScatRec})~\cite{mattausch2017image} : This is an inverse problem based approach is referred here as ScatRec~\cite{mattausch2017image}, which reconstructs scatterer map based on a single observation.
\item Deep Learning based estimation of parametric scatterer maps ({\bf ScatParam}) : This is our proposed method trained on simulated images, as detailed in the previous section.
\end{itemize}
For deconvolution based methods, TRF and ScatRec, in simulated data we used the known PSF from the simulations and for acquired data, we used a cepstrum-domain PSF estimation method described in~\cite{mattausch2016image}, followed by least square fitting to the known parametric form in Eq.~(\ref{eq:psf}) to project them on our PSF model manifold, which our trained network is better conditioned on.

Since different scatterer representations cannot be compared directly and no ground truth is available, we evaluated the performance of scatterer estimation on the envelope images simulated from the estimated scatterer maps. 
For re-synthesizing images, we used the same forward simulation for all the methods, namely a discrete image-space convolution of the scatterers estimated by any particular method with the same depth-dependent PSFs estimated for deconvolution based methods. 
The convolution was implemented in Matlab to operate separately for each PSF on rows of image pixels.

\subsection{Evaluation Metrics}
Several evaluation metrics are utilized to assess the simulation performance and compare our method ScatParam with SampleEnv, TRF, and ScatRec. 
Three image-based metrics, mean image intensity (I), signal-to-noise ratio (SNR), contrast-to-noise ratio (CNR) and one histogram based metric, Kullback-Leibler (KL) divergence, are used to calculate the mismatch between ground truth and simulated envelope images. 

\textbf{Mean Image Intensity (I)} can capture any global intensity shift in the simulated images, which can be caused by systematic biases in constructive or destructive interference when the scatterers are not distributed truly stochastically or in a view-dependent way.
For instance, if the scatterers estimated from one direction align in a structured way, when the object is imaged from an oblique direction (e.g., with the half-wavelength projected on the rotation angle aliased with the structure), these scatterers may then interfere with each other mostly destructively, creating an artificial intensity drop in the image. 
The change in mean image intensity is calculated as follows
\begin{equation}
    \Delta \textrm{I} = \frac{|\textrm{I}_t-\textrm{I}_s|}{\textrm{I}_t}, \quad \textrm{with} \quad \textrm{I} = \frac{1}{N} \sum_{j=1...N} s_j,
\end{equation}
where $s_j$ denotes the image intensity value at the $j$-th pixel, and
$N$ is the number of pixels.
$\textrm{I}_t$ and $\textrm{I}_s$ are the mean image intensities of ground truth and simulated image, respectively. 
Hereafter, the subscript $t$ refers to the ground truth and $s$ to a simulated image.

\textbf{Signal-to-noise Ratio (SNR)} measures the global statistics of signal-and-noise ratio. 
For the simulation purpose, we aim at reproducing the images at the same SNR level as the ground truth. 
Any mismatch in SNR is then quantified as follows
\begin{equation}
    \Delta \textrm{SNR} = \frac{|\textrm{SNR}_t-\textrm{SNR}_s|}{\textrm{SNR}_t}, \quad \textrm{with} \quad \textrm{SNR} = \frac{\mu}{\sigma},
\end{equation}
where $\mu$ and $\sigma$ denote the mean and standard deviation of envelope image intensities.

\textbf{Contrast-to-noise ratio (CNR)} mismatch is defined similarly as 
\begin{equation}
    \Delta \textrm{CNR} = \frac{|\textrm{CNR}_t-\textrm{CNR}_s|}{\textrm{CNR}_t}, \quad \textrm{with} \quad \textrm{CNR} = \frac{|\mu_{s1}-\mu_{s2}|}{\sigma_{s1}+\sigma_{s2}},
\end{equation}
where $\mu_{s1}$, $\mu_{s2}$, $\sigma_{s1}$, and $\sigma_{s2}$ denote the means and standard deviations of the image intensities within two contrasting regions.
This metric is clinically relevant, as incorrect tissue contrast in the simulated images may lead to the learning of false diagnostic cues during medical training.

\textbf{Kullback-Leibler (KL) divergence} compares the statistics between the histograms of two images as follows:
\begin{equation}
    \textrm{KL} (h_s || h_t) = \sum_{l=1...D} h_s[l] \log\left(\frac{h_s[l]}{h_t[l]}\right),
\end{equation}
where $h_t$ and $h_s$ are the normalized histograms of the true and simulated images respectively. The number of histogram bins $D$ is set herein to $50$.
Histogram statistics are widely explored for tissue characterization~\cite{shankar1993use, tsui2008classification}. 
Hence, a large discrepancy in the histograms could indicate a mismatch in the speckle pattern appearance.
Since computing histograms over the whole image could miss local speckle texture information, we calculate a KL divergence metric locally within patches (herein non-overlapping patches of $3\times3\,\textrm{mm}^2$ corresponding to 10\,$\lambda$ per dimension) and report herein the metric mean over all patches.

For calculating $\Delta$SNR, $\Delta$CNR and KL divergence, simulated image ($s$) is normalized (or brightness equalized) with respect to ground truth image ($t$) by multiplying a factor $\frac{\sum_j t_j}{\sum_j s_j}$ similarly to~\cite{mattausch2017image}, to eliminate effects in these metrics from any global intensity shift, which is captured separately by $\Delta \textrm{I}$.

\subsection{Synthetic Data}
This experiment evaluates the invariance of the reconstructed scatterer maps to various US imaging parameters.
To that end, we used simulated images from Field\,II with controllable imaging conditions. 
A numerical phantom of $15\times15\,\textrm{mm}^2$ with a 3\,mm circular inclusion was simulated for imaging at 6.0\,MHz center frequency, with a 128 element 40\,mm linear transducer, a single transmit focus at the center, and dynamic receive focusing. 
The phantom is placed 15\,mm away from the transducer to avoid near field effects. 

We evaluate the simulation results for 
1) phantom rotation, which emulates imaging a region of interest from different viewing directions;
2) phantom compression, which emulates image plausibility under potential tissue deformation, e.g. induced by probe compression.

\subsubsection{Rotation Experiment}
\begin{figure}
\includegraphics[width=1\linewidth]{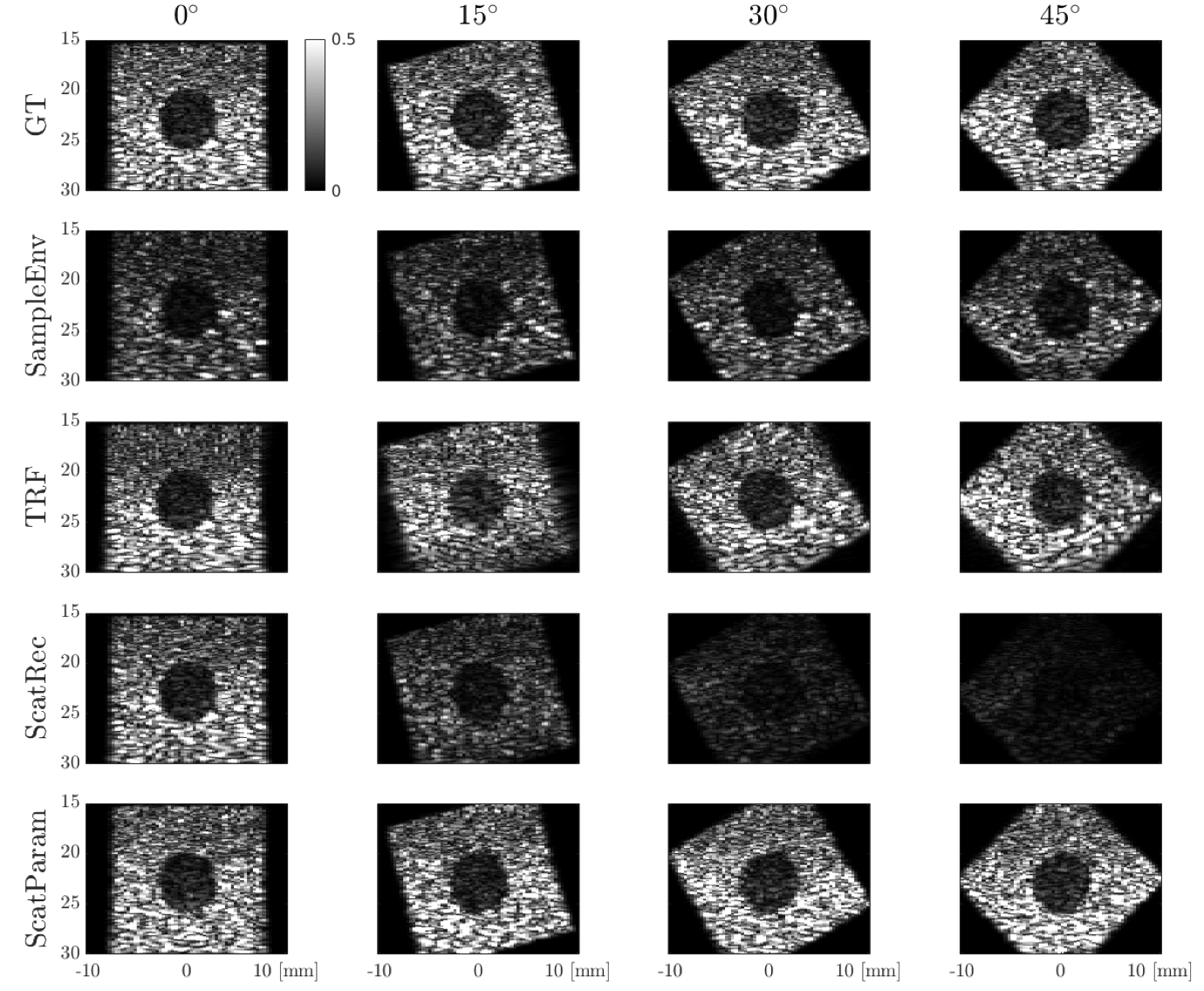}
\caption{Simulated images of the numerical phantom for views rotated by $15^{\circ}$, $30^{\circ}$, and $45^{\circ}$.}
\label{fig:rotation_inclusion}
\end{figure}

In this experiment, we evaluate the invariance of the reconstructed scatterer maps to phantom rotation.
For this, the box phantom were rotated around the phantom center with varying angles.
Fig.~\ref{fig:rotation_inclusion} illustrates the results for the rotated views.
The images are cropped to the region of interest.
The top row shows the ground truth envelope images simulated by Field II for varying rotation angles ranging from $0^{\circ}$ to $45^{\circ}$ with a $15^{\circ}$ increment.
For ScatRec, the image intensity drastically decreases with increasing rotation angle and the phantom almost disappears for $45^{\circ}$ due to destructive interference, as reported in~\cite{mattausch2017image}.
The images simulated by SampleEnv appear dark, whereas the image mean intensity remains similar to the ground truth images for TRF and ScatParam during rotation. 
It can be well observed that the reconstruction of the proposed ScatParam is robust for different viewing angles.

\begin{figure}
\includegraphics[width=1\linewidth]{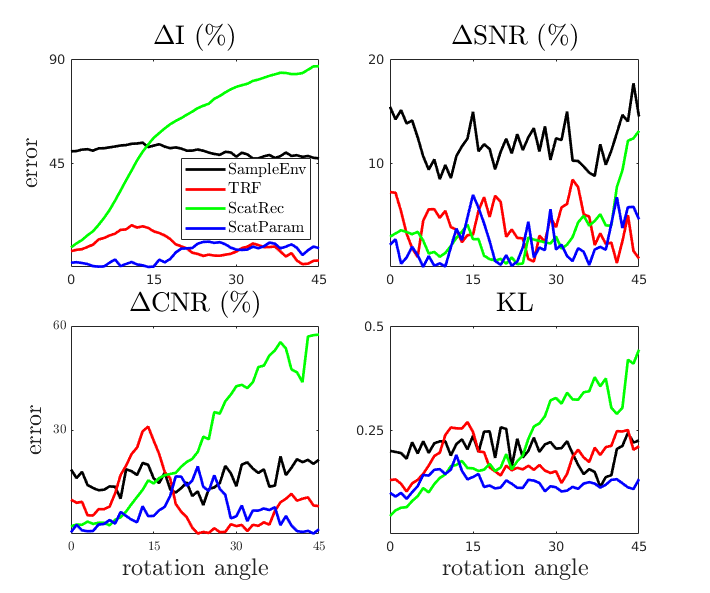}
\caption{Performance with respect to rotation angle in terms of normalize mean image intensity difference ($\Delta \textrm{I}$ (\%)), signal-to-noise ratio difference ($\Delta \textrm{SNR}$ (\%)), contrast-to-noise ratio difference ($\Delta \textrm{CNR}$ (\%)) and histogram difference (KL divergence).}
\label{fig:rotation_inclusion_err_curves}
\end{figure}

We evaluate the simulation performance quantitatively for the above experiment by investigating the error for rotations with $1^\circ$ increments, with the results plotted in Fig.~\ref{fig:rotation_inclusion_err_curves}.
ScatRec is close to the ground truth near $0^\circ$ given any metric, but it deviates largely from the ground truth at larger rotation angles, which corroborates the visual observations above.
The errors for SampleEnv are consistently high irrespective of rotation, indicating not a successful scatterer reconstruction.
In general, the error metrics of TRF vary without any pattern, remaining relatively similar across the angles, except a large variance in CNR.
This is in agreement with the observation in Fig.~\ref{fig:rotation_inclusion} that the tissue contrast in the $15^\circ$ rotated view of TRF is diminished. 
The metrics for ScatParam exhibit minor fluctuations and remain overall relatively low and thus superior compared to the other methods.

Tab~\ref{tab:quant_results_rotation_circluar}. summarizes the mean, median and maximum error across the $1^\circ$ increment results in Fig.~\ref{fig:rotation_inclusion_err_curves}. 
ScatParam is seen to achieve the lowest error overall, nearly $30 \%$ lower than the second best method in $\Delta \textrm{CNR}$ and KL divergence, demonstrating the representativeness and robustness of the estimated scatterer map with respect to probe rotation. 
\begin{table}
\setlength{\tabcolsep}{2pt}
\caption{Mean (mean), median (med), and maximum (max) errors for rotations between $0^\circ$ and $45^\circ$ with $1^\circ$ increments. Bold number indicates the smallest value per column. }
\label{tab:quant_results_rotation_circluar}
\centering
\scalebox{0.9}{
   \begin{tabular}{c|rrr|rrr|rrr|rrr}
   Metric  & \multicolumn{3}{c|}{$\Delta \textrm{I}$ (\%)} & \multicolumn{3}{c|}{$\Delta \textrm{SNR}$ (\%)} & \multicolumn{3}{c|}{$\Delta \textrm{CNR}$ (\%)} & \multicolumn{3}{c}{KL $(\times10^{-2})$} \\
   \hline
   \rowcolor{Gray}
    &mean &med &max &mean &med &max &mean &med &max &mean &med &max \\
   \hline
   SampleEnv &50.1 &50.2 &50.8 & 11.9 & 11.7 & 17.7 & 16.3 & 16.6 & 22.4 & 20.4 & 20.7 & 25.5 \\
  \hline
  TRF & 9.1 & 8.6 & 18.1 &4.0 &3.7 &8.4 & 9.4 & 8.0 & 31.0 &16.7 &16.5 &24.1\\
  \hline
  ScatRec & 59.6 & 68.0 & 86.9 &3.4 &2.8 &13.1 &27.2 &22.7 &57.5 &22.2 &18.2 &44.2 \\
  \hline
  ScatParam & \bf{5.7} & \bf{7.4} & \bf{11.0} &\bf{2.3} &\bf{1.8} &\bf{6.9} & \bf{6.6} &\bf{5.2} &\bf{19.5} &\bf{12.5} &\bf{12.3} &\bf{19.3} 
\end{tabular}}
\end{table}

\subsubsection{Compression Experiment}
\begin{figure}
\includegraphics[width=1\linewidth]{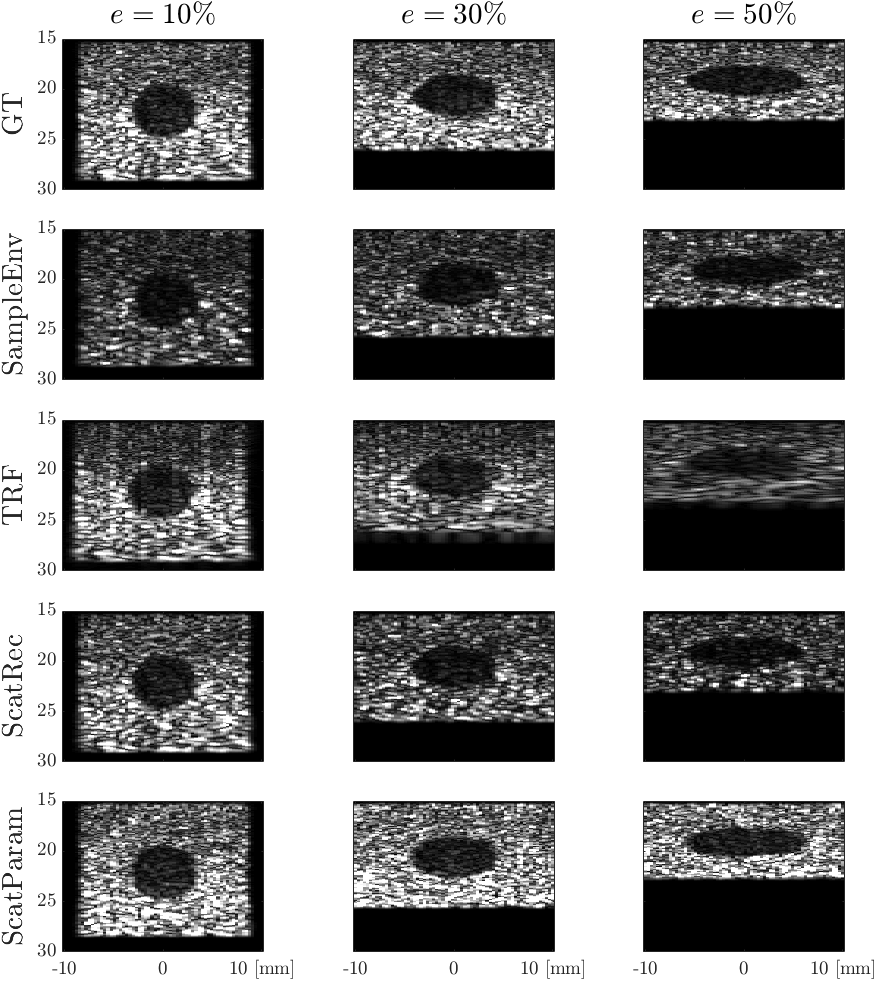}
\caption{Simulated images of the numerical phantom for axial compressions of 10\%, 30\%, and 50\% strain.}
\label{fig:compression_inclusion}
\end{figure}

With this, we investigate the invariance of reconstructed scatterer maps with respect to physical deformation.
An axial phantom compression was simulated by interpolating the estimated scatterer maps on grids deformed by varying levels of axial strain $e$.
Simulated images with all methods at $e=\{10,30,50\}\%$ compression are shown in Fig.~\ref{fig:compression_inclusion}. 
The images simulated by TRF are corrupted by aliasing artifacts. 
The simulation results of SampleEnv and ScatRec look similar with reduced intensity and degraded speckle pattern for large compression strain. 
The simulated images by ScatParam are visually closest to the ground truth images in terms of speckle appearance and contrast, but slightly hyperechoic for large compression. 

\begin{figure}
\includegraphics[width=1\linewidth]{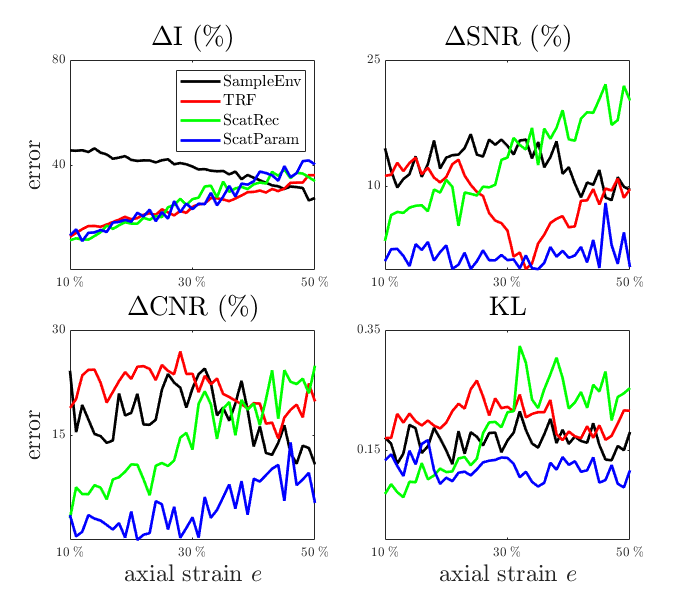}
\caption{Performance evolution with increasing axial strain $e$ in terms of difference in image mean intensity ($\Delta \textrm{I}$~(\%)), signal-to-noise ratio ($\Delta \textrm{SNR}$~(\%)), contrast-to-noise ratio ($\Delta \textrm{CNR}$~(\%)) and histogram difference (KL).}
\label{fig:compression_inclusion_err_curves}
\end{figure}

We investigate the simulation errors for compression ranging from $e=10$\% to $e=50$\% with $1$\% increments in Fig.~\ref{fig:compression_inclusion_err_curves}. 
Similarly to the rotation experiment, the error metrics of ScatRec increase proportionally with increasing compression, whereas our method performs consistently superior.
SampleEnv and TRF yield large errors, especially in $\Delta \textrm{CNR}$ and KL divergence metrics.

The observations are supported by numerical results shown in Tab.~\ref{tab:quant_results_compression_circular}, reporting the mean, minimum, and maximum errors for the above plots.
Our proposed method ScatParam is seen to be superior in terms of $\Delta$SNR, $\Delta$CNR, and KL divergence. 
It achieves approximately $78 \%$ lower error in the mean $\Delta \textrm{SNR}$, $68 \%$ in the mean $\Delta \textrm{CNR}$ and $28 \%$ in the mean KL divergence compared to the second best method.
TRF achieves $6 \%$ lower error than ScatParam in the mean $\Delta$I here.
\begin{table}
\setlength{\tabcolsep}{2pt}
\caption{Mean (mean), median (med) and maximum (max) errors across 10\% to 50\% strain with 1\% increments. Bold number indicates the smallest value per column.}
\label{tab:quant_results_compression_circular}
\centering
\scalebox{0.9}{
   \begin{tabular}{c|rrr|rrr|rrr|rrr}
  Metric  & \multicolumn{3}{c|}{$\Delta \textrm{I}$ (\%)} & \multicolumn{3}{c|}{$\Delta \textrm{SNR}$ (\%)} & \multicolumn{3}{c|}{$\Delta \textrm{CNR}$ (\%)} & \multicolumn{3}{c}{KL $(\times10^{-2})$} \\
   \hline
   \rowcolor{Gray}
    &mean &med &max &mean &med &max &mean &med &max &mean &med &max \\
   \hline
   SampleEnv &38.5 &39.4 &46.3 & 12.7 & 13.3 & 16.2 &17.6 & 17.2 & 24.5 & 16.6 & 16.5 & 21.4 \\
  \hline
  TRF & \bf{24.4} & \bf{24.1} & \bf{36.1} &8.3 &9.3 &13.3 & 21.5 & 22.2 & 26.9 &20.3 &20.4 &26.6 \\
  \hline
  ScatRec & 25.6 & 27.1 & 38.3 &12.9 &13.1 &22.1 & 14.6 & 14.6 & 24.9 & 18.9 &19.9 &32.3 \\
  \hline
  ScatParam & 26.0 & 25.1 & 41.7 &\bf{1.8} &\bf{1.6} &\bf{8.0} & \bf{4.7} &\bf{4.1} &\bf{13.9} &\bf{11.9} & \bf{11.7} &\bf{16.6} 
\end{tabular}}
\end{table}

\subsection{Gelatin Phantom}
\begin{figure*}
\centering
\minipage{0.7\textwidth}
 \centering
 \includegraphics[width=1\linewidth]{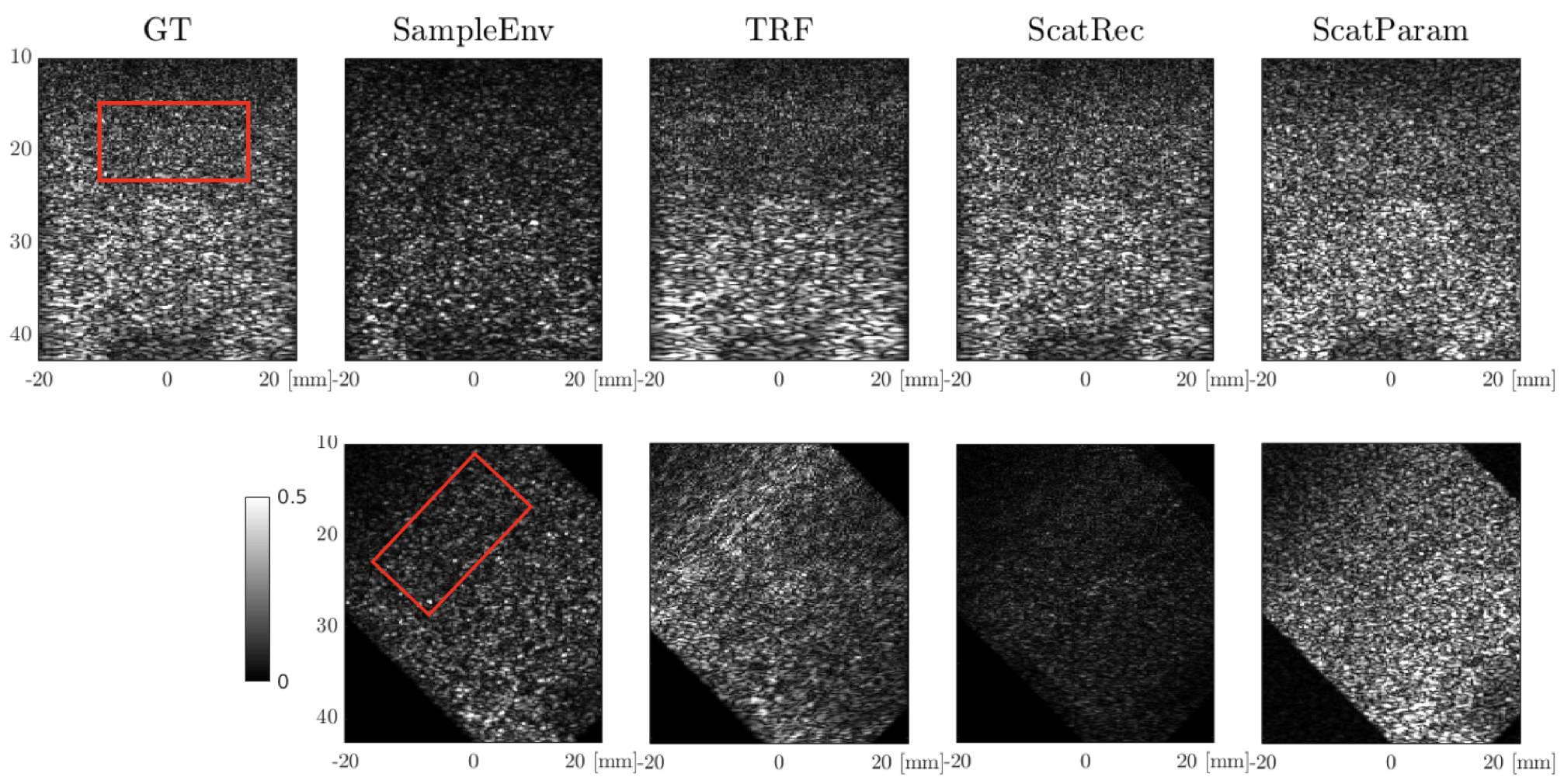}
 \caption*{(a)}
\endminipage
\hfill
\minipage{0.3\textwidth}
 \centering
 \includegraphics[width=1\linewidth]{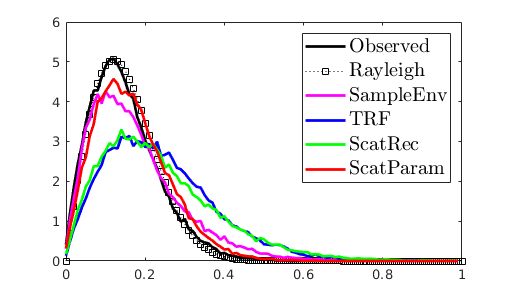}
 \caption*{(b)}
\endminipage
\caption{(a) With scatterers estimated from the gelatin phantom, images simulated at $0^{\circ}$ (top) and $45^{\circ}$ (top). (b)~Histogram of the ground truth and rotated views inside the area shown by a red rectangle in one sample image. Quantifying the difference between the ground truth and simulated image histograms using KL divergence indicates to the following errors: $0.101$ for SampleEnv, $0.593$ for TRF, $0.588$ for ScatRec, and $0.032$ for ScatParam.}
\label{fig:gelatin_results}
\end{figure*}

Next we investigate the performance of our method for a real ultrasound scan of a gelatin phantom with corn starch as the scattering medium.
A circular inclusion is made by adding twice as high as the concentration of starch in the background. 
The beamformed RF images are collected by a Fukuda Denshi UF-760AG ultrasound machine with a linear probe FUT-LA385-12P.
The results in the top row of Fig.~\ref{fig:gelatin_results}(a) show an excellent agreement of ScatRec with the ground truth image, as expected from an overconstrained optimization, while the SampleEnv and TRF results exhibiting speckle textures different than the ground truth.
Our proposed method ScatParam generates images almost indistinguishable from the observation in terms of speckle texture and tissue contrast.
The experiment is further evaluated with quantitative metrics in Tab.~\ref{tab:gelatin_phatom}.
\begin{table}
\setlength{\tabcolsep}{2pt}
\caption{Quantitative evaluation metrics for the gelatin phantom. Bold number indicates the smallest error per column.}
\label{tab:gelatin_phatom}
\centering
\begin{tabular}{c|r|r|r|r}
\textbf{Method} & $\Delta \textrm{I}$ (\%) & $\Delta \textrm{SNR}$ (\%) & $\Delta \textrm{CNR}$ (\%) &KL ($\times 10^{-2}$) \\ \hline
SampleEnv        & 44.0 & 19.2 & 13.3  & 25.5 \\
\hline
TRF              & 5.0 & 10.0 & 24.0 & 21.2  \\
\hline
ScatRec          &\bf{1.7}  & \bf{0.6} & \bf{1.3} & \bf{2.7}\\
\hline
ScatParam        & 7.4  & 0.9  & 8.1 & 17.0 
\end{tabular}
\end{table}
ScatRec achieves the closest match to the ground truth image, as its scatterer estimation overfits to the observed phantom image.
Comparing the remaining three, ScatParam yields CNR, SNR, and KL divergence metrics lower than TRF and SampleEnv, indicating that our method better preserves contrast and intensity distribution.

The bottom row in Fig.~\ref{fig:gelatin_results}(a) shows the simulated $45^{\circ}$ rotated views of the phantom.
For an isotropic scattering phantom, the mean echo and speckle texture statistics would be invariant to the viewing direction and hence, after rotation, a speckle appearance similar to the initial view is expected. 
Such coherence pattern may naturally not stay pixel-wise constant after, e.g., rotation, therefore no pixel-wise error metric was employed.
Similarly to the rotation experiment of Field II, the image simulated by ScatRec appears hypoechoic for $45^{\circ}$ angle, while the speckle pattern of TRF is severely distorted.

The ground truth rotated views are not obtained here.
Nevertheless, we evaluated the results by comparing the histogram of the observed $0^{\circ}$ envelope image to the histograms of the rotated envelope images (after brightness equalization) for the homogeneous region inside the red rectangle depicted in Fig.~\ref{fig:gelatin_results}(a).
Here Rayleigh statistics serve as an important criterion for assessing ultrasound speckle texture, as the envelope intensity should follow a Rayleigh distribution for fully developed speckles~\cite{goodman1975statistical}.
The histogram of ScatEnv closely follows the observed ground truth histogram, i.e.\ an ideal Rayleigh distribution, consistent with the observation reported in~\cite{alessandrini2015pipeline}.
Nevertheless, our approach ScatParam is seen to have an even better agreement with the observed pattern; indeed, with an over 3-folds lower KL divergence score as listed in the figure caption.

\subsection{In Vivo Experiment}
\begin{figure*}
\centering
\includegraphics[width=1\linewidth]{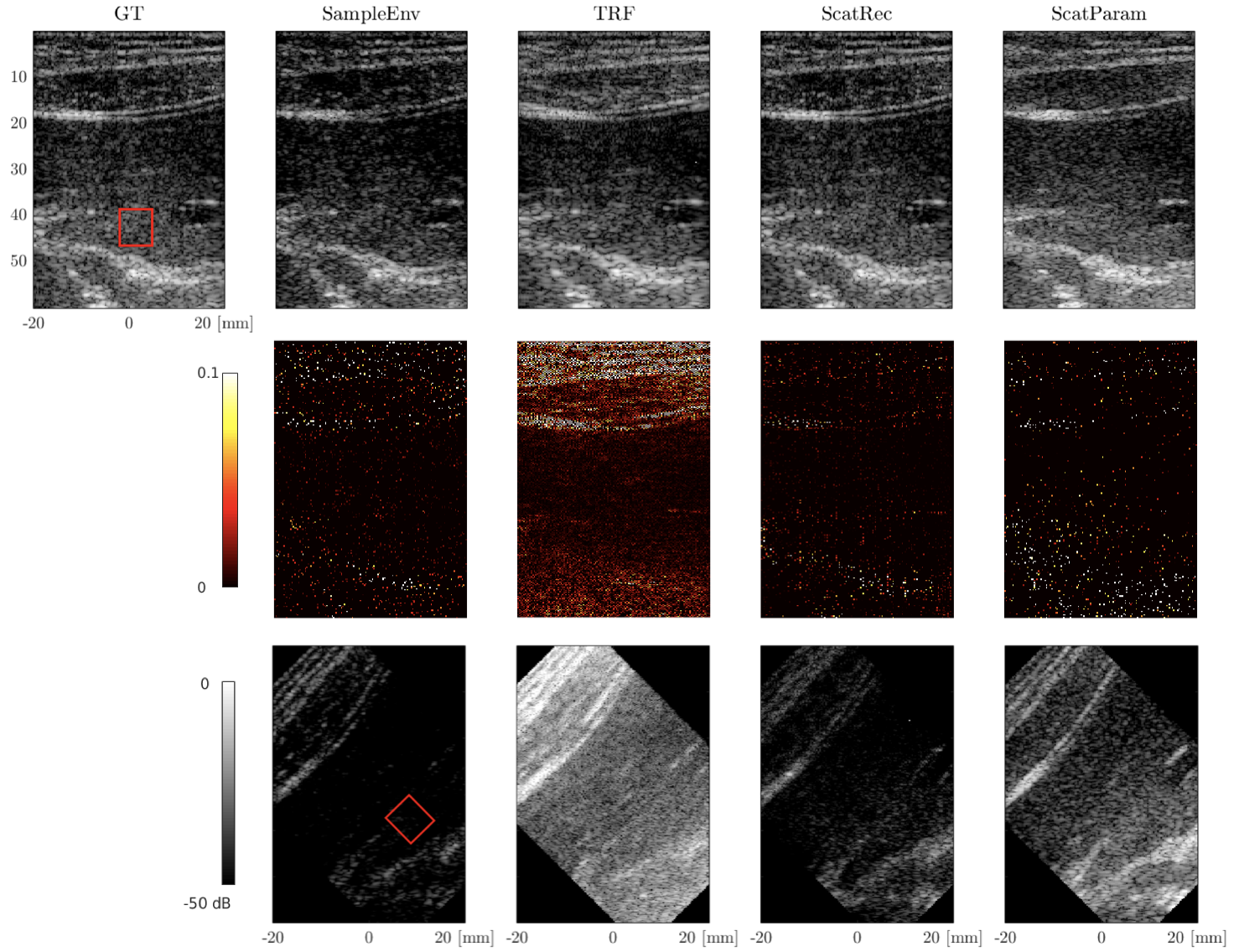}
\caption{Images simulated in the original $0^\circ$ configuration from scatterers estimated from an in vivo liver image~(top), estimated scatterer maps, with only a random 8\% shown for sake of visualization~(middle), and the images simulated from the same scatterers after a rotation of $45^\circ$~(bottom). B-mode images are shown here, in contrast to all other images in this paper showing envelope images.}
\label{fig:invivo_images}
\end{figure*}

For this purpose, beamformed RF data from an in-vivo scan of the liver was collected.
Scatterer representations were estimated for the imaged region using all four methods presented.
Fig.~\ref{fig:invivo_images} depicts the results simulated in the acquisition configuration of $0^\circ$ and simulating a rotation of $45^\circ$. 
At $0^\circ$, all methods perform somewhat similarly, with SampleEnv slightly hypoechoic, TRF misrepresenting speckle texture, and ScatRec reconstructing an exact replica, as expected.
Noticeably, given the dynamic compression for B-mode images presented for the in-vivo experiment, the loss of brightness in the image generated by ScatEnv is less prominent compared to the previous experiments showing envelope images.
At $45^\circ$, however, SampleEnv and ScatRec both become very hypoechoic and preserve very little structural detail, and TRF becomes overly hyperechoic with the speckle pattern almost disappeared -- indicating that these three all did not estimate robust scatterer representations.
In contrast, our method ScatParam emulates the image rotation realistically despite the change in PSF, preserving mean intensity as well as structural detail and contrast, after rotation.

\begin{figure}
\centering
\minipage{0.45\textwidth}
 \centering
 \includegraphics[width=1\linewidth]{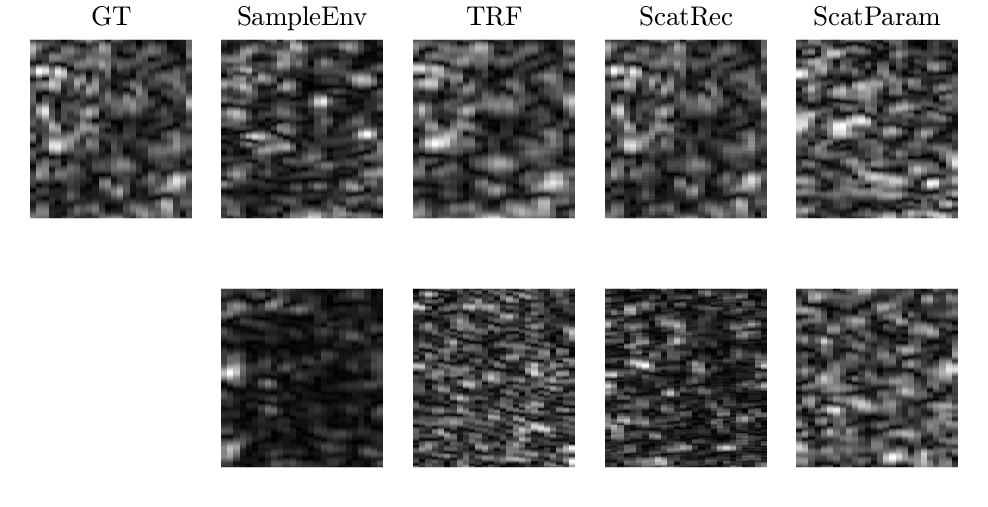}
 \caption*{(a)}
\endminipage
\hfill
\minipage{0.3\textwidth}
 \centering
 \includegraphics[width=1\linewidth]{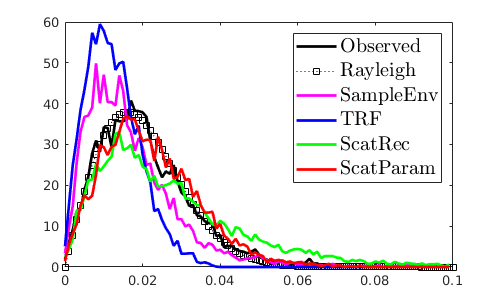}
 \caption*{(b)}
\endminipage
\caption{(a) Image patches in the red marked region in Fig.~\ref{fig:invivo_images}(a) for the $0^\circ$ view (top) and the $45^\circ$ view (bottom). Envelope images after brightness equalization are shown here. (b) Histogram of the ground truth patch compared with the simulated rotated patches, showing a best match with ScatParam. 
Quantifying the difference between the ground truth and simulated image histograms using KL divergence indicates to the following errors: $0.040$ for SampleEnv, $0.059$ for TRF, $0.018$ for ScatRec, and $0.004$ for ScatParam.}
\label{fig:invivo_rotation}
\end{figure}

Assuming a homogeneous isotropic structure for the liver tissue (marked with a red rectangule in Fig.~\ref{fig:invivo_images}), a histogram comparison is also performed to compare post-rotation speckle appearance to the original image.
Given the clear difference in mean intensities, a raw B-mode histogram comparison indicates the superiority of ScatParam, by a large margin.
For a comparison of the speckle texture alone, we therefore first brightness equalized the images (cf.\ Fig.~\ref{fig:invivo_rotation}(a)) and then performed the histogram comparison.
As seen in Fig.~\ref{fig:invivo_rotation}(b), ScatParam histogram even after after rotation closely matches to that of original ground truth image histogram, following an ideal Rayleigh distribution.
KL divergence results reported in Fig.~\ref{fig:invivo_rotation} caption corroborate with this observation.

\section{Discussion}
In this work, we demonstrated a learning based approach for probabilistic scatterer estimation in the context of realistic ultrasound image simulation.
The proposed framework ScatParam involves sparse scatterer model with Gaussian distribution for scatterer amplitude and estimation of the Gaussian mean directly from US envelope images by neural network.
Similar isotropic scatterer distributions were used in several earlier works, e.g.~\cite{jensen1996field, alessandrini2011restoration, burger2012real, mattausch2017image, alessandrini2015pipeline,  bamber1980ultrasonic}.
In our preliminary experiments for estimating parameter maps of both mean and standard deviation, we have found that different such combinations may generate similar image outputs, making such double parameter estimation ill-posed.  
Therefore, in this work we fixed the standard deviation and estimated only the mean value.
In addition, we assumed a fixed density value for our model, since it is shown that in case of fully developed speckles the tissue characteristics are not affected by density~\cite{thijssen1990texture}. 
Estimating scatterer density in addition to the amplitude mean could allow us to distinguish between partially and fully developed speckles; nevertheless, we observed comparable performance with only mean estimation for both numerical simulations and in-vivo experiments presented herein. 

For complex in-vivo scatterer distributions more sophisticated models may be required.  
For example, for muscle fibers, anisotropic parametrizations such as with tensor, wavelet, or frequency-domain representations may be more suitable. Indeed, instead of hand-crafted parametric models, (arbitrary) distributions could potentially be parametrized using a neural network to be inferred from observed images.

In the paper, we have conducted several experiments evaluating the invariance of tissue properties with geometric transformations, which is of great importance for ensuring plausible simulation of the same tissue content under different imaging conditions. 
Rotation and axial compression are chosen, since they well represent the clinical examinations such as transducer in-plane tilting and compression. 
Furthermore, these experimental scenarios model potential variation in speckle appearance from directional changes (isotropy) and concentration changes (axial strain) in scatterer configurations.

In comparison to other deconvolution algorithms, our method does not require PSF estimation in advance.
It was reported in~\cite{mattausch2017image} that accurate PSF estimation is needed as input to an inverse problem based method, analogously important for the Wiener filter and other deconvolution algorithms.
Our trained CNN is able to capture the PSF information in the form of different speckle textures in input images, hence estimated parameter maps would be independent of imaging system.
One can produce visually plausible images without considering PSF as in the method SampleEnv.
However, the simulated speckle statistics with SampleEnv herein are not fully in agreement with the observations, since by dismissing the modulated nature of PSF and thereby any destructive interference, SampleEnv cannot fully model the interference between scatterers. 
Setting scatterer amplitudes directly using envelope intensities causes the incorrect translation of speckle variations into scatterer maps, i.e. higher amplitude scatterers lumped around the peaks of speckles. This leads to brighter hyperechoic and darker hypoechoic regions even after PSF convolution. Such granular appearance is less visible after dynamic range compression in B-mode images, e.g. in Fig. 8, whereas these and resulting overall intensity reduction become apparent in the envelope images shown for the rest of the experimental results. Note that we present envelope images, as they allow easier interpretation of speckle patterns and imaging physics, where effects of methodological choices on the results are not masked by any graymap transform.

We herein chose Wiener filtering as a basic deconvolution baseline for TRF, with low computation and memory requirements. The results with artifacts and reduced contrast of TRF indicate that the use of the same low resolution of the input RF image in the output scatterers hinder interpolation in the scatterer domain after transformations, such as compressions. Although there are more recent forms of TRF, e.g. ~\cite{florea2018axially, besson2019physical}, ScatRec~\cite{mattausch2017image} was chosen herein as a state-of-the-art baseline performing a sophisticated deconvolution approach as an optimization of an inverse-problem definition; with a suitable model, i.e. norms and regularizers; with depth-dependent PSF; non-negative scatterer constraint; and with a higher solution scatterer resolution than the RF image domain. Regardless of baselines, our results indicate that our learning based solution ScatParam as a fast-implementable network solution performs satisfactorily for image simulation and without requiring complex convolution modeling, PSF estimation, iterative optimization, or any other complex processing steps.
Note that using only Gaussian noise model for the training samples, ScatParam is able to successfully estimate scatterer parametrezations for no-noise numerical experiments as well as unknown-noise phantom and in-vivo examples, potentially indicating the robustness of the proposed parametrization and the respectively trained NN to an assumed noise model.

Underconstrained inverse problem based approaches can be improved by using multiple measurements of the same tissue with different imaging parameters. 
A successful example was illustrated in~\cite{mattausch2017image}, where ScatRec with multiple observations from beam steering is shown to be more robust to viewing angle changes. 
However, the aligned beam-steered images are in general not available from clinical scanners, let alone raw RF data.
Our method takes envelope images as input and can thus accept clinical B-Mode images with slight modification of network training.
For a fair comparison, we herein used the single view version of inverse problem, i.e.\ ScatRec1 in~\cite{mattausch2017image}.

In this work any ultrasound image appearance is attributed solely to isotropic scattering, where attenuation variations and coherent reflections are not considered.
For instance, all the methods compared in Fig.~\ref{fig:gelatin_results}(a) attribute the slight attenuation in the original image behind the gelatin inclusion (after a depth of 40\,mm) to some form of lower scattering amplitude, thereby resulting in the attenuation not correctly reproduced in the rotated images, e.g.\ extending diagonally rather than vertically. This demonstrates the need to take directional attenuation and reflections into account during any scatterer estimation process, which should be a focus of future studies.
Furthermore, any potential reflections at anatomical boundaries, i.e. between supra-wavelength structures, would also be attributed to result from the reconstructed scatterers, which may then incorrectly reproduce the tissue from different viewing angles. 
In all our numerical examples, any such reflections were thus avoided.

The gelatin phantom was made with isotropic scatterers and for the in-vivo example the liver was chosen for its relatively homogeneous speckle appearance. Phantom and in-vivo evaluations were conducted within small regions-of-interest selected in line with our assumptions.
But, for instance in the presented ultrasound acquisitions, the directional reflection from the isoechoic inclusion in the gelatin phantom and reflections at muscle boundaries in the in-vivo liver image appear similarly after rotation, which is suboptimal as reflection effects should be direction-dependent.
Nevertheless, it may be possible to separate the directional dependent image content prior to capturing only scattering effects as described herein. 
One can then simulate such directional wave interactions at a later time using ray tracing techniques~\cite{burger2012real, mattausch2018realistic}.
For instance, reflection boundaries could be detected and removed using a simple phase symmetry (PS) algorithm~\cite{hacihaliloglu2009bone}, which is designed to estimate directional reflections at tissue boundaries such as bone surfaces, as thin hairline structures. In a way similar to estimating and compensating for reflections, acoustic attenuation can indeed also be reconstructed a-priori~\cite{rau2019attenuation} in order to spatially normalize incident acoustic energy to decouple its effect from our reconstructed scatterer amplitudes.

Our network training takes approximately 6 hours on Nvidia Titan XP GPU.
Once trained, our network can estimate scatterer maps in milliseconds at inference time, for arbitrary input image size (as being a fully convolutional network architecture).
In contrast, the inverse problem based approach takes approximately 2 hours for one image from a single view. 
For multiple (beam-steered) observations, the computation time would increase exponentially, quickly making this method infeasible for large images and 3D volumes. 

\section{Conclusion}
We have demonstrated a learning-based technique to efficiently estimate the distribution of tissue scatterer representation, which can then be fed directly into convolution- or ray-tracing-based simulation techniques~\cite{burger2012real, mattausch2018realistic} to simulate realistic images for sonographer training. 
The proposed network is trained only with synthetic images generated with random shapes and spatial invariant convolution. 
In comparison to the state-of-the-art methods, we demonstrate with numerical simulations the proposed estimation pipeline being robust for simulating images at different viewing angles and tissue deformations.
The method is further evaluated on a tissue-mimicking gelatin phantom and an in-vivo liver image, demonstrating the generalization ability of our network to real ultrasound scans.

\ifCLASSOPTIONcaptionsoff
  \newpage
\fi

\bibliographystyle{IEEEtran}
\bibliography{references}
\end{document}